\documentclass[11pt]{article}

\usepackage[authoryear]{natbib}

\usepackage[margin=1in]{geometry}
\usepackage{amsmath,amssymb,mathtools,bm}
\usepackage{booktabs}
\usepackage{microtype}
\usepackage{hyperref}
\usepackage[nameinlink,noabbrev]{cleveref}
\usepackage{tikz}
\usetikzlibrary{arrows.meta,positioning,fit,backgrounds}

\newcommand{\calG}{\mathcal G}

\newcommand{\ind}{\mathbb I}
\newcommand{\E}{\mathbb E}

\newcommand{\sat}{\sigma}
\newcommand{\rmin}{r_{\min}}

\newcommand{\eat}[1]{}

\title{A note on goal-based hierarchical RL}
\author{Kevin Murphy}
\date{\today}

\begin{document}
\maketitle

\begin{abstract}
  The agent-centric general value function (ACGVF) construction
of \citet{tasse2026goal}
lets the agent make two decisions  that are normally
imposed by the environment or agent designer:
which goal to pursue and when to declare a goal as finished
(in addition to choosing the action).
This is a very general framework that subsumes almost all prior
work on reinforcement learning, control and planning,
as well as more general formalisms proposed in the cognitive sciences.
However, it assumes the environment is fully observed,
i.e., that  the observation is a sufficient statistic.
In \citet{murphy2025rl},
a general agent design was proposed where the policy
is based on an internal belief state $z_t$ 
and an internal goal;
however, the goals were assumed to be externally provided.
In this note, we unify and extend these two approaches using
the formalism of hierarchical hidden Markov models (HHMM)
\citep{murphy2001hhmm}.
\eat{
The top level state
$Q_t^2=G_t$ is the agent's current goal,
the lower state $Q_t^1=Z_t$ is its
within-goal internal state, and the lower-level finish variable $F_t^1=D_t$
is the agent's done decision.  The defining HHMM constraint says that $G_t$
is held fixed while $D_t=0$ and may be replaced by an agent-selected goal only
when $D_t=1$.  This gives a precise probabilistic semantics to a continual
agent that alternates between selecting goals and pursuing them under partial
observability.
}
\end{abstract}

\section{Introduction}

  The agent-centric general value function (ACGVF) construction
of \citet{tasse2026goal}
starts with a fully observed MDP and augments the agent's
decision with a goal $G_t\in\calG$ and a Boolean done action
$D_t\in\{0,1\}$.  Its key conceptual step is that goals
and termination are selected by the agent in the service of environmental
reward, rather than specified as part of the task.  However, it assumes
the world state $S_t$ is used directly to define the policy
$\pi(g,a,d\mid s)$.  Thus $S_t$ is implicitly both the Markov state of the
world and the sufficient information state of the controller.

The generic agent architecture in~\citet{murphy2025rl} separates these two
objects,
and uses a policy of the form
$\pi_\theta(a\mid z,g)$, where $g$ is the goal
and $Z_t$ is the internal memory state
updated according to
\begin{equation}
 Z_{t+1}=U_\phi(Z_t,A_t,O_{t+1}).
 \label{eq:memory-update}
\end{equation}
The policy therefore acts on the history summary $Z_t$, not directly on the
observation $O_t$.  However, it treats $g$ as an external input.

In this note, we propose to combine these approaches
by making the goal another component of the agent's
dynamical state.  There are now two internal timescales:
\begin{enumerate}
  \item the \emph{goal state} $G_t$, which is piecewise constant; and
  \item the \emph{within-goal state} $Z_t$, which is updated after every
  observation.
\end{enumerate}
The done action is the boundary event coupling them.  This is exactly the
call--return semantics represented by the Hierarchical Hidden Markov Model
formalism introduced in \citet{murphy2001hhmm}.

\section{A controlled two-level HHMM}

Let $X_t$ denote the hidden physical state of the environment and $O_t$ its
observation.  A convenient one-step timing convention is
\begin{equation}
 (Z_t,G_t)\ \longrightarrow\ (A_t,D_t)\ \longrightarrow\
 (X_{t+1},O_{t+1},R_{t+1})\ \longrightarrow\ Z_{t+1}\
 \longrightarrow\ G_{t+1}.
 \label{eq:timing}
\end{equation}
Thus $D_t$ says whether the transition produced by $A_t$ completes the
current goal.  

For a trajectory of length $T$, the joint distribution can be factored as
\begin{align}
 &p(x_0,o_0,z_0,g_0)
 \prod_{t=0}^{T-1}
 \pi_\theta(a_t,d_t\mid z_t,g_t)\,
 p_{\mathrm{env}}(x_{t+1},o_{t+1},r_{t+1}\mid x_t,a_t)
 \nonumber\\[-1mm]
 &\hspace{36mm}\times
 q_\phi(z_{t+1}\mid z_t,a_t,o_{t+1})\,
 p_\mu(g_{t+1}\mid
 g_t,z_{t+1},d_t).
 \label{eq:joint}
\end{align}
Note that this joint distribution is defined by
the coupling of the agent
components $\pi_{\theta}$, $q_{\phi}$ and $p_{\mu}$ --- which generate
$Z_t$, $G_t$, $D_t$, $A_t$ ---
and the environment's $p_{\mathrm{env}}$,
which generates its own hidden state $X_t$ (not seen by agent),
$O_t$ and $R_t$.
More precisely, the components are as follows:
\begin{itemize}
  \item $\pi_{\theta}$ is the goal-conditioned policy over low-level actions $a_t$
and the decision to stop the low level, $d_t$.

\item $q_{\phi}$ is the agent state update function.
For a deterministic recurrent state, $q_\phi$ is a point mass at the output
of an update map $U_\phi$.  For a belief-state or latent world-model agent,
$q_\phi$ can instead be a stochastic filtering distribution.

\item  $p_{\mu}$ is the induced distribution over goals, defined as follows:
\begin{equation}
 p_\mu(g_{t+1}\mid g_t,z_{t+1},d_t)
 =
 \begin{cases}
   \delta_{g_t}(g_{t+1}), & d_t=0,\\
   \mu_\eta(g_{t+1}\mid z_{t+1}), & d_t=1.
 \end{cases}
 \label{eq:goal-gate}
\end{equation}
The goal selector $\mu_\eta$ is a learned upper-level policy.  It is evaluated
only at a boundary; between boundaries it degenerates to the identity map.
This is stronger than merely feeding a freshly sampled goal to a policy on
every step: it gives the selected goal temporal persistence
(thus goals become like options or macro actions; \citealp{Sutton99}).
\end{itemize}

\begin{figure}[t]
\centering
\begin{tikzpicture}[
  x=1cm,y=1cm,
  var/.style={circle,draw,minimum size=9mm,inner sep=1.3pt,fill=white},
  obs/.style={circle,draw,minimum size=9mm,inner sep=1.3pt,fill=black!18},
  arr/.style={-{Latex[length=2.1mm,width=1.4mm]},semithick}
]
\foreach \t/\x in {1/0,2/3.8,3/7.6} {
  \node[var] (qh\t) at (\x,4.3)        {$Q^2_{\t}$};   
  \node[var] (f\t)  at (\x+1.75,2.55)  {$F^1_{\t}$};   
  \node[var] (ql\t) at (\x,0.9)        {$Q^1_{\t}$};   
  \node[obs] (y\t)  at (\x,-1.15)      {$Y_{\t}$};     
}

\foreach \t in {1,2,3} {
  \draw[arr] (qh\t) -- (ql\t);                          
  \draw[arr] (qh\t) -- (f\t);                           
  \draw[arr] (ql\t) to[out=70,in=235] (f\t);
  \draw[arr] (ql\t) -- (y\t);                           
  \draw[arr] (qh\t) to[out=-55,in=35,looseness=0.95] (y\t);
}

\foreach \t/\u in {1/2,2/3} {
  \draw[arr] (qh\t) -- (qh\u);                          
  \draw[arr] (ql\t) -- (ql\u);                          
  \draw[arr] (f\t)  -- (qh\u);                          
  \draw[arr] (f\t)  -- (ql\u);                          
}
\end{tikzpicture}
\caption{A two-level HHMM drawn as a DBN, adapted from
  \citet{murphy2001hhmm}.
  (Note:  we use superscript $1$ to denote
the \emph{lower} (concrete) level and superscript $2$ the \emph{upper}
(abstract) level,  following convention of \citet{theocharous2004hpomdp},
which is the reverse of the convention in  \citet{murphy2001hhmm}.)
  At time $t$ the system is in state
$(Q^2_t,Q^1_t)$, and the binary variable $F^1_t$ turns on iff the current
lower-level sub-model has just finished; $F^1_t$ does not affect $Y_t$, so a
sub-process boundary is never directly observed and must be inferred.  Note the
ordering implied by the arcs into slice $t+1$: the new abstract state
$Q^2_{t+1}$ is drawn first (from $Q^2_t$ and $F^1_t$), and only then is the new
concrete state $Q^1_{t+1}$ drawn, from $Q^1_t$, $F^1_t$ and the freshly chosen
parent $Q^2_{t+1}$.  When $F^1_t=0$ both levels persist horizontally; when
$F^1_t=1$ the parent may move and the child is re-entered.}
\label{fig:hhmm-dbn}
\end{figure}

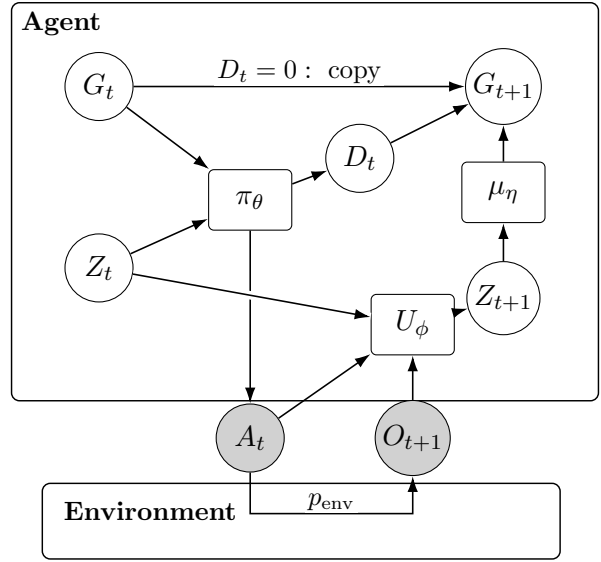
\begin{figure}[t]
\centering
\begin{tikzpicture}[
  x=1cm,y=1cm,
  var/.style={circle,draw,minimum size=9mm,inner sep=1.3pt,fill=white},
  interface/.style={circle,draw,minimum size=9mm,inner sep=1.3pt,fill=black!18},
  mechanism/.style={rectangle,draw,rounded corners=2pt,minimum width=11mm,
    minimum height=8mm,inner sep=2pt,fill=white},
  arr/.style={-{Latex[length=2.1mm,width=1.4mm]},semithick},
  box/.style={draw,rounded corners=3pt,line width=.55pt},
  label/.style={font=\small\bfseries,anchor=north west},
  note/.style={font=\small,fill=white,inner sep=1pt}
]
\node[var]       (gt) at (0,4.2) {$G_t$};
\node[var]       (zt) at (0,1.8) {$Z_t$};
\node[mechanism] (pi) at (2.0,2.7) {$\pi_\theta$};
\node[var]       (dt) at (3.45,3.25) {$D_t$};
\node[mechanism] (update) at (4.15,1.05) {$U_\phi$};
\node[var]       (zn) at (5.35,1.4) {$Z_{t+1}$};
\node[mechanism] (mu) at (5.35,2.8) {$\mu_\eta$};
\node[var]       (gn) at (5.35,4.2) {$G_{t+1}$};

\node[interface] (at) at (2.0,-0.45) {$A_t$};
\node[interface] (on) at (4.15,-0.45) {$O_{t+1}$};

\draw[arr] (gt) -- (pi);
\draw[arr] (zt) -- (pi);
\draw[arr] (pi) -- (dt);
\draw[arr] (dt) -- (gn);
\draw[arr] (gt) -- node[note,above] {$D_t=0:\ \text{copy}$} (gn);
\draw[arr] (zn) -- (mu);
\draw[arr] (mu) -- (gn);

\draw[arr] (pi) -- (at);
\draw[arr,preaction={draw=white,line width=3.2pt,-}] (zt) -- (update);
\draw[arr] (at) -- (update);
\draw[arr] (on) -- (update);
\draw[arr] (update) -- (zn);

\coordinate (ein) at (2.0,-1.45);
\coordinate (eout) at (4.15,-1.45);
\draw[arr] (at) -- (ein) -- node[note,above] {$p_{\mathrm{env}}$} (eout) -- (on);

\begin{scope}[on background layer]
  \node[box,fit=(gt)(zt)(pi)(dt)(update)(zn)(mu)(gn),
    inner xsep=7mm,inner ysep=6mm] (agentbox) {};
  \draw[box] (-0.75,-2.05) rectangle (6.10,-1.05);
\end{scope}
\node[label] at (agentbox.north west) {Agent};
\node[label] at (-0.60,-1.15) {Environment};
\end{tikzpicture}
\caption{A two-slice factor graph representation
  of a two level conditional HHMM representing a hierarchical
  agent policy $\pi_{\theta}$ with goal selection function
  $\mu_{\eta}$ and internal state update   function $U_{\phi}$.  Shading marks
the observed interface variables $A_t$ and $O_{t+1}$; all nodes inside the
Agent box are hidden internal variables or mechanisms.  The Environment box
maps the action to the next observation (its internal state and reward are
omitted).  When $D_t=0$ (done node is false),
the goal $G_t$ is copied forward; when $D_t=1$, the agent
selects a new goal using $\mu_\eta(G_{t+1} \mid Z_{t+1})$.
The reward generated from the environment is assumed to be part of $O_{t+1}$.
}
\label{fig:dbn}
\end{figure}

We now discuss how to represent this distribution as a hierarchical
hidden Markov model.
A generic two-level HHMM, drawn as a DBN, is shown in \cref{fig:hhmm-dbn}.
Following that figure we use superscript $1$ for the lower (concrete) level and
superscript $2$ for the upper (abstract) level.
The correspondence with our agent
variables is
\begin{equation}
 Q_t^2=G_t,\qquad Q_t^1=Z_t,\qquad F_t^1=D_t.
 \label{eq:mapping}
\end{equation}
Here $G_t$ is the abstract (goal) state, and $Z_t$ are the low level states
that generate the sequence of actions.
We assume  $D_t=0$ (meaning the agent isn't done with its current goal)
forces $G_{t+1}=G_t$, whereas $D_t=1$ returns control to
the top level  and permits a horizontal transition to a new goal.  For a genuinely
continual agent the top-level finish variable $F_t^2$ can be clamped to zero;
it is needed only if the entire interaction can terminate. 

For the agent, the outputs $Y_t$ of this generative process
are its actions $A_t$. Furthermore, the whole process is conditioned
on the environment's observations $O_t$.
We can thus represent our modified hierarchical controller
as the factor graph
shown in \cref{fig:dbn}.

Note that we make an important change to the original HHMM formulation
from \citet{Fine98}, where entering a new high level state $Q_t^2$
reinitializes its child state $Q_t^1$ from a top-down prior.
A general agent should not erase its belief about the world
whenever it changes goals,
so we allow $Z_{t+1}$ to always depend on $Z_t$.  Given that $Z_{t+1}$ is
already determined by the time a boundary is reached, the natural ordering
within a slice is to choose the next goal $G_{t+1}$ given $Z_{t+1}$, rather
than the other way round; this inverts the vertical arc of
\cref{fig:hhmm-dbn}.
The underlying reason is a change of role: we are using the HHMM as an
observation-conditioned generator of actions, so $Z_t$ is driven by the
incoming observations and must persist across goals.  In the hierarchical POMDP
of \citet{theocharous2004hpomdp} it is instead an action-conditioned generator
of observations, and the lower state is a sub-model index that may legitimately
be reset on entry.
(This is related to the notion of cross-segment boundaries
in a segmental HMM \citep{Ostendorf96}.)

\eat{
\paragraph{An important qualification.}
A literal HHMM \citep{Fine98}
normally reinitializes its child state when the parent calls a
new submodel.  A general agent should not erase its belief about the world
whenever it changes goals.  There are two clean solutions.  The minimal one is
to let the boundary kernel in~\cref{eq:joint} preserve the relevant components
of $Z_t$.  The stricter graphical-model solution decomposes
\begin{equation}
 Z_t=(B_t,M_t),
 \label{eq:belief-phase}
\end{equation}
where $B_t$ is persistent task-independent memory (ideally a belief or
predictive state) and $M_t$ is a goal-local controller phase.  A goal boundary
resets $M_t$ conditional on $(B_t,G_{t+1})$ but carries $B_t$ forward,
as in a segmental HMM \citep{Ostendorf96}.
Under
this decomposition, the most faithful HHMM identification is
$Q_t^2=G_t$ and $Q_t^1=M_t$, with $B_t$ an observed input to the hierarchy.
Writing $Q_t^1=Z_t$ remains valid for a generalized, input--output HHMM whose
vertical transition has access to the persistent memory.
}

\section{Goal satisfaction and agent-centric reward in latent state}

Under partial observability, determining
goal satisfaction (which is needed to compute the reward)
cannot in general be a
predicate of the inaccessible world transition $(s,a,s')$,
as was assumed in  \citep{tasse2026goal}.
Instead we replace it with
an internal satisfaction model
\begin{equation}
\sigma_t = \sat_\psi(g_t \mid z_t,a_t,o_{t+1},z_{t+1})\in[0,1].
 \label{eq:satisfaction}
\end{equation}
This may describe a target set in latent space, a proposition inferred from
observations, or a distributional target.  The internal reward becomes
\begin{align}
 \bar r_{t+1}
 &=\bigl[(1-d_t)+d_t \sigma_t \bigr]r_{t+1}
 +d_t\bigl[1- \sigma_t \bigr]\rmin.
 \label{eq:internal-reward}
\end{align}
Consequently, continuing receives ordinary environmental reward, correct
termination receives environmental reward, and incorrect termination receives
the lower-bound penalty $\rmin$.  This is the latent-state analogue of 
Eq. 2 in \citep{tasse2026goal}.

For a fixed goal, define the within-goal action value
and value function as 
\begin{align}
 Q_g(z,a,d)
 &=\E\!\left[
 \bar R_{t+1}
 +(1-d)V_g(Z_{t+1})
 \mid Z_t=z,G_t=g,A_t=a,D_t=d
 \right],
 \label{eq:acgvf-bellman}\\
 V_g(z)&=\max_{a,d}Q_g(z,a,d).
 \label{eq:goal-value}
\end{align}
This recovers the ACGVF Bellman equation after replacing $s$ by the agent's
information state $z$.  The greedy lower controller
$\pi_g^*(a,d\mid z)$ provides \emph{mastery}: for a fixed $g$, it tries both
to reach the goal and to recognize when it has done so.  The upper controller
then chooses among mastered goals:
\begin{equation}
 \mu^*(g\mid z)>0
 \quad\Longrightarrow\quad
 g\in\arg\max_{g'\in\calG}V_{g'}(z).
 \label{eq:goal-selection}
\end{equation}
Thus goal selection, via $\mu^*(g \mid z)$,
and goal achievement (mastery), via $\pi^*_g(a,d \mid z)$,
are related but distinct problems.

\section{Segment values versus continual values}

Equation~\eqref{eq:acgvf-bellman} assigns zero continuation value after
$d=1$.  It therefore values one goal-directed segment at a time.  The HHMM
view exposes a useful extension when ``done'' ends a goal but does not reset
the physical world.  Let $W_g(z)$ be the value while pursuing $g$ and
$H(z)=\max_g W_g(z)$ be the value at an upper-level decision point
after the goal is finished.
(By contrast, $V_g(z)$ is the value of state $z$
assuming that achieving goal $g$ ends the process.)
A discounted
continual recursion is
\begin{align}
 W_g(z)
 =\max_{a,d}\E\!\left[
   \bar R_{t+1}
   +\gamma\{(1-d)W_g(Z_{t+1})+dH(Z_{t+1})\}
   \mid z,g,a,d
 \right],
 \label{eq:continual-bellman}
\end{align}
The $d H(Z_{t+1})$ term is precisely the return from the lower HHMM to the
upper controller.  Removing this term (and setting $\gamma=1)$
yields the segment-level $V_g(z)$.  Keeping
it optimizes a stream of successive, agent-delimited goals.
Correspondingly, the goal-selection rule of~\cref{eq:goal-selection} becomes
$\mu^*(g\mid z)>0\Rightarrow g\in\arg\max_{g'}W_{g'}(z)$, whose value is
exactly $H(z)$; the segment-level form in~\cref{eq:goal-selection} is recovered
by dropping the continuation term.

To see the difference more clearly,
suppose two actions complete the current goal with the same
immediate reward.  One leaves the agent near many valuable future goals; the
other leaves it trapped in an unproductive state.  $V_g$ may regard them as
equivalent, while $W_g$ prefers the first because it accounts for what happens
after goal completion.
In other words,  we have
\begin{equation}
 \begin{array}{c|c}
 V_g(z) & d=1 \ \Longrightarrow\ \text{future value }0 \\
 W_g(z) & d=1 \ \Longrightarrow\ \text{future value }H(Z_{t+1})
 \end{array}
\end{equation}

At goal boundaries,~\cref{eq:continual-bellman} induces the usual semi-Markov
upper-level return.  If a goal selected at boundary time $\tau_k$ terminates at
$\tau_{k+1}$, its macro-transition is
\begin{equation}
 p(z_{\tau_{k+1}},\,\mathcal R_k,\,\Delta_k
 \mid z_{\tau_k},g_k),
 \quad
 \mathcal R_k=\sum_{j=0}^{\Delta_k-1}\gamma^j\bar r_{\tau_k+j+1},
 \quad
 \Delta_k=\tau_{k+1}-\tau_k.
 \label{eq:smdp}
\end{equation}
Hence the construction is simultaneously (i) an HHMM in primitive time and
(ii) an option-like semi-MDP \citep{Sutton99}
at goal boundaries.  Its distinctive feature is
that the option label, termination decision, and latent information state are
all internal to the agent.

\eat{
\section{When is \texorpdfstring{$Z_t$}{Zt} a sufficient statistic?}

The construction does not require the environment's observation $O_t$ to be
generated by a Markov process.
However, it
does require the controller state to be sufficient for predicting quantities
relevant to control.  An exact Bayesian choice is the belief state
\begin{equation}
Z_t = p(X_t= \cdot \mid O_{0:t},A_{0:t-1}),
 \label{eq:belief}
\end{equation}
possibly augmented with the current goal.  More generally, a learned $Z_t$
is adequate when histories mapped to the same $z$ agree, to the accuracy
needed by the policy, on
\begin{enumerate}
  \item the distribution of future observations and rewards under candidate
  actions;
  \item the probability that each candidate goal is satisfied; and
  \item the continuation values of the available goals.
\end{enumerate}
This is a control-sufficiency rather than reconstruction requirement.  A
world-model loss helps learning when rewards are sparse,
but predicting every pixel is not necessary,
so long as the features it does generate are sufficient to
determine reward or goal
satisfaction.
}

\section{Learning and inference}

There are two interpretations of the hidden nodes, leading to different
algorithms.

\paragraph{Goals are latent explanations.}
If $G_t$ and $D_t$ are not recorded---for example, when discovering goals from
demonstrations---then~\cref{eq:joint} is a controlled HHMM likelihood.
In this case we can estimate parameters using EM.
The DBN construction of~\citet{murphy2001hhmm} gives an algorithm
for an exact E step for discrete states that is  linear in sequence
length for fixed hierarchy depth and state cardinality.  Amortized filtering,
particle methods, or structured variational inference are natural for neural
and continuous $Z_t$.  The persistence constraint in~\cref{eq:goal-gate}
prevents the goal label from changing arbitrarily at every step and therefore
acts as a useful segmentation prior.

\paragraph{Goals are internal actions.}
If the agent chooses its own values for $G_t$ and $D_t$, then these variables
are observed from the agent's perspective even though they are hidden from the
environment.  One can learn:
\begin{itemize}
  \item $U_\phi$ jointly with the policies using the actor--critic gradients
  described below;
  \item $Q_g$ or $W_g$ using TD learning on tuples
  $(z_t,g_t,a_t,d_t,r_{t+1},o_{t+1},z_{t+1})$;
  \item $\pi_\theta(a,d\mid z,g)$ using value-based or actor--critic updates;
  and
  \item $\mu_\eta(g\mid z)$ using only boundary transitions, with the
  duration-aware return in~\cref{eq:smdp}.
\end{itemize}
The single primitive-time replay stream can train both levels; boundaries
provide the segmentation automatically.
We give more details below.

\paragraph{Joint objective.}
A generic actor--critic objective is
\begin{equation}
 \mathcal L
 =\lambda_{\mathrm{sat}}\mathcal L_{\mathrm{sat}}(\psi)
 +\lambda_{\mathrm{TD}}\mathcal L_{\mathrm{TD}}(\omega)
 +\lambda_{\mathrm{crit}}\bigl[
    \mathcal L_{\mathrm{critic}}(\nu)+\mathcal L_{\mathrm{critic}}(\kappa)
  \bigr]
 -J(\theta,\eta,\phi)
 +\lambda_{\mathrm{dur}}\Omega_{\mathrm{dur}}(D).
 \label{eq:objective}
\end{equation}
Here $\mathcal L$ is the total objective to be minimized, and all expectations
in its component terms are taken over replay data or trajectories generated by
the agent.  The components are:
\begin{itemize}
\item $\mathcal L_{\mathrm{sat}}(\psi)$ trains the satisfaction model
  $\sat_\psi$ in~\cref{eq:satisfaction} to predict whether goal $g$ is
  satisfied; $\psi$ denotes its parameters.
  If an independent hard or soft satisfaction target
  $\widetilde Y^{\mathrm{sat}}_{t+1}\in[0,1]$ is available, a concrete choice
  is the binary cross-entropy
\begin{align}
 \mathcal L_{\mathrm{sat}}(\psi)
 =-\E_{\mathcal D_{\mathrm{replay}}}\!\left[
   \widetilde Y^{\mathrm{sat}}_{t+1}\log\sigma_t
   +(1-\widetilde Y^{\mathrm{sat}}_{t+1})\log(1-\sigma_t)
 \right],
 \label{eq:satisfaction-loss}
\end{align}
where $\sigma_t$ is defined in~\cref{eq:satisfaction}.  Under partial
observability, satisfaction is not generally observed, so this supervised loss
is applicable only when an independent target can be obtained from an
environment label, an observable certificate, hindsight, or a belief-state
model.  For example, let
$b_{t:t+1}(x,x')=p(X_t=x,X_{t+1}=x'\mid O_{0:t+1},A_{0:t})$ be the filtered
two-slice belief.  If the goal predicate is known on world-state transitions,
a soft target is
\begin{equation}
 \widetilde Y^{\mathrm{sat}}_{t+1}
 =\sum_{x,x'}b_{t:t+1}(x,x')
   \Pr\!\left((x,a_t,x')\models g_t\right).
 \label{eq:belief-satisfaction-target}
\end{equation}
For the simpler goal ``the next state lies in $g_t$,'' this reduces to
$\widetilde Y^{\mathrm{sat}}_{t+1}
=\sum_{x'}B_{t+1}(x')\ind\{x'\in g_t\}$.  If no independent source of targets
exists, satisfaction must instead be treated as latent and learned through a
weakly supervised or latent-variable objective.  Using $\sigma_t$ as its own
target would provide no learning signal.

\item $\mathcal L_{\mathrm{TD}}(\omega)$ is a temporal-difference loss, such as a
  squared Bellman residual, for the segment value $Q_g$ or the continual value
  $W_g$; $\omega$ denotes the critic parameters.
A concrete value-based definition of the TD loss is
\begin{equation}
 \mathcal L_{\mathrm{TD}}(\omega)
 =\E_{\mathcal D_{\mathrm{replay}}}\!\left[
   \left(Q_{\omega,g_t}(z_t,a_t,d_t)-y_t\right)^2
 \right],
 \label{eq:td-loss}
\end{equation}
where $\mathcal D_{\mathrm{replay}}$ is the replay distribution.  For the
segment and continual formulations, respectively, suitable targets are
\begin{align}
 y_t^{\mathrm{seg}}
 &=\bar r_{t+1}+(1-d_t)V_{\omega^-,g_t}(z_{t+1}),
 \label{eq:segment-td-target}\\
 y_t^{\mathrm{cont}}
 &=\bar r_{t+1}+\gamma\left[
   (1-d_t)W_{\omega^-,g_t}(z_{t+1})
   +d_tH_{\omega^-}(z_{t+1})
 \right],
 \label{eq:continual-td-target}
\end{align}
where $\omega^-$ denotes fixed target-network parameters and
$H_{\omega^-}(z)=\max_g W_{\omega^-,g}(z)$.

\item $\mathcal L_{\mathrm{critic}}(\nu)$ and $\mathcal
  L_{\mathrm{critic}}(\kappa)$ are the regression losses that fit the
  actor--critic baselines used to estimate the policy gradient of $J$; they are
  defined in~\cref{eq:lower-critic,eq:upper-critic} below.  These are distinct
  from $\mathcal L_{\mathrm{TD}}(\omega)$: the latter learns the optimal
  internal-reward values $Q_g$ or $W_g$, whereas the former learn on-policy
  values of the environmental return.  Together with $-J$ these form the
  actor--critic pair, but they are weighted separately: $J$ carries an implicit
  coefficient of $1$ and thus sets the scale against which every other term is
  balanced, while $\lambda_{\mathrm{crit}}$ controls how strongly the critic
  regression competes with it (and, through $Z_t^\phi$, how much it shapes the
  shared state representation).

\item $J(\theta,\eta,\phi)$ is the expected environmental return obtained by the
  lower policy $\pi_\theta(a,d\mid z,g)$ and the upper goal-selection policy
  $\mu_\eta(g\mid z)$.  The minus sign appears because $J$ is maximized while
  $\mathcal L$ is minimized.  The dependence on $\phi$ is explicit because the
  policies act on the learned state $Z_t$.
  The policy objective can be defined by
\begin{equation}
 J(\theta,\eta,\phi)
 =\E_{\tau\sim p_{\theta,\eta,\phi}}\!\left[
   \sum_{t=0}^{\infty}\gamma^t R_{t+1}
 \right],
 \label{eq:policy-return}
\end{equation}
where $p_{\theta,\eta,\phi}$ is the trajectory distribution induced by the
lower policy, goal selector, state-update model, and environment.  Omitting
$\phi$ from $J$ is justified only when the representation is fixed or gradients
from the actor objective are stopped at $Z_t$.  Thus the TD
loss uses the internal reward $\bar R$, including the incorrect-termination
penalty, to learn goal mastery, whereas $J$ uses the environmental reward $R$
to ground goal selection in external utility.  If incorrect termination
should also be penalized directly in the actor objective, $R_{t+1}$ can be
replaced by $\bar R_{t+1}$ in~\cref{eq:policy-return}; in that case $J$ also
depends on $\psi$ through the satisfaction model.

  \item $\Omega_{\mathrm{dur}}(D)$ regularizes the sequence of done
  decisions $D=(D_0,D_1,\ldots)$, and hence the induced distribution of segment
  durations $\Delta_k$.
The duration regularizer is optional but practically useful: without an
inductive bias, $D_t\equiv1$ yields one-step goals and $D_t\equiv0$ yields no
boundaries.  A deliberation cost, minimum-duration constraint  (c.f., \citep{Dong2020}),
termination
entropy, or prior over segment lengths can prevent these degenerate
factorizations.
Similarly, if the goal vocabulary is learned, diversity or
information constraints may be needed to avoid duplicate or unreachable goal
labels.  These regularizers define the abstraction; environmental reward then
decides which abstractions are useful.

  \item $\lambda_{\mathrm{sat}},\lambda_{\mathrm{TD}},\lambda_{\mathrm{crit}}$,
  and $\lambda_{\mathrm{dur}}$ are nonnegative
  hyperparameters controlling the relative contribution of the corresponding
  terms.
\end{itemize}

\paragraph{Optimization.}
For any parameter block $v\in\{\phi,\psi,\omega,\nu,\kappa,\theta,\eta\}$, the
gradient of the joint minimization objective is
\begin{align}
 \nabla_v\mathcal L
 &=\lambda_{\mathrm{sat}}\nabla_v\mathcal L_{\mathrm{sat}}
  +\lambda_{\mathrm{TD}}\nabla_v\mathcal L_{\mathrm{TD}}
  +\lambda_{\mathrm{crit}}\nabla_v\bigl[
     \mathcal L_{\mathrm{critic}}(\nu)+\mathcal L_{\mathrm{critic}}(\kappa)
   \bigr]
  -\nabla_v J
  +\lambda_{\mathrm{dur}}\nabla_v\Omega_{\mathrm{dur}}.
 \label{eq:joint-gradient}
\end{align}
A term is zero when it does not depend on $v$ or when its gradient has been
explicitly stopped.  The relative scales and directions
of the component gradients can nevertheless cause interference, which can be
managed using the $\lambda$ weights, gradient normalization, or separate
optimizer steps.

Because the action, done, and goal choices are discrete and the environment is
not normally differentiable, the gradients of~\cref{eq:policy-return} are
usually estimated with policy-gradient or actor--critic methods.  To define
the required advantages, let
\begin{equation}
 \mathcal U_t=\sum_{j=0}^{\infty}\gamma^jR_{t+j+1}
 \label{eq:environment-return-to-go}
\end{equation}
be the environmental return-to-go appearing in $J$.  Writing the quantities
below as functions of $(z,g)$ presupposes that the pair $(Z_t,G_t)$ is a
sufficient statistic for control, i.e.\ that histories mapped to the same
$(z,g)$ agree on the distribution of future observations and rewards under
candidate actions.  This is exact when $Z_t$ is the belief state
$p(X_t=\cdot\mid O_{0:t},A_{0:t-1})$, and is the operative approximation when
$Z_t$ is learned.  Under that assumption the lower-level values
under the current hierarchical policy are
\begin{align}
 Q^{\mathrm{low}}(z,g,a,d)
 &=\E[\mathcal U_t\mid Z_t=z,G_t=g,A_t=a,D_t=d],
 \label{eq:lower-q}\\
 V^{\mathrm{low}}(z,g)
 &=\E_{(A,D)\sim\pi_\theta(\cdot\mid z,g)}
   [Q^{\mathrm{low}}(z,g,A,D)].
 \label{eq:lower-v}
\end{align}
At a boundary time $\tau_k$, the corresponding upper-level values are
\begin{align}
 Q^{\mathrm{upper}}(z,g)
 &=\E[\mathcal U_{\tau_k}\mid Z_{\tau_k}=z,G_{\tau_k}=g],
 \label{eq:upper-q}\\
 V^{\mathrm{upper}}(z)
 &=\E_{G\sim\mu_\eta(\cdot\mid z)}[Q^{\mathrm{upper}}(z,G)].
 \label{eq:upper-v}
\end{align}
These are policy-evaluation values for the environmental objective $J$, rather
than the optimal internal-reward values $Q_g$, $V_g$, and $W_g$ defined
earlier.  (If the actor objective is switched to the internal reward, as
discussed below, then $\omega$ and $\nu$ end up optimizing and evaluating the
same reward function, but they remain distinct: $Q_g$ and $W_g$ are optimal
values, whereas $V^{\mathrm{low}}$ and $V^{\mathrm{upper}}$ are on-policy
values for the current $(\theta,\eta)$.)

We write $V_\nu^{\mathrm{low}}(z,g)$ and $V_\kappa^{\mathrm{upper}}(z)$ for
learned critics, with parameters $\nu$ and $\kappa$, approximating
\cref{eq:lower-v,eq:upper-v}.  At the upper level it is also convenient to name
the environmental reward accumulated over one goal segment,
\begin{equation}
 \mathcal R_k^{\mathrm{env}}
 =\sum_{j=0}^{\Delta_k-1}\gamma^j R_{\tau_k+j+1},
 \label{eq:upper-environment-return}
\end{equation}
the environmental-reward counterpart of $\mathcal R_k$ in~\cref{eq:smdp}.

\paragraph{Fitting the critics.}
The critics are fit by regression onto bootstrapped targets, at the two
timescales of the hierarchy.  The lower critic is updated at every primitive
step,
\begin{align}
 \mathcal L_{\mathrm{critic}}(\nu)
 &=\E\!\left[\bigl(y_t^{\mathrm{low}}
    -V_\nu^{\mathrm{low}}(Z_t,G_t)\bigr)^2\right],
 &
 y_t^{\mathrm{low}}
 &=R_{t+1}+\gamma V_{\nu^-}^{\mathrm{low}}(Z_{t+1},G_{t+1}),
 \label{eq:lower-critic}
\end{align}
and the upper critic only at the boundary times $\tau_k$,
\begin{align}
 \mathcal L_{\mathrm{critic}}(\kappa)
 &=\E\!\left[\bigl(y_k^{\mathrm{upper}}
    -V_\kappa^{\mathrm{upper}}(Z_{\tau_k})\bigr)^2\right],
 &
 y_k^{\mathrm{upper}}
 &=\mathcal R_k^{\mathrm{env}}+\gamma^{\Delta_k}
   V_{\kappa^-}^{\mathrm{upper}}(Z_{\tau_{k+1}}),
 \label{eq:upper-critic}
\end{align}
where $\nu^-$ and $\kappa^-$ denote target (stop-gradient) copies of the
parameters; without them the regression chases its own output.  The factor
$\gamma^{\Delta_k}$ in~\cref{eq:upper-critic} makes the upper target
duration-aware, as in the semi-Markov transition of~\cref{eq:smdp}.  Note that
$G_{t+1}$ in~\cref{eq:lower-critic} is the \emph{gated} goal
of~\cref{eq:goal-gate}: it equals $G_t$ when $D_t=0$, and is a fresh draw from
$\mu_\eta(\cdot\mid Z_{t+1})$ when $D_t=1$.  The lower-level bootstrap
therefore absorbs a goal switch automatically, and no special case is needed at
a boundary.

\paragraph{The upper critic is redundant.}
Comparing~\cref{eq:lower-v} with~\cref{eq:upper-q} shows that
$Q^{\mathrm{upper}}$ and $V^{\mathrm{low}}$ are the same function of $(z,g)$:
both equal $\E[\mathcal U_t\mid Z_t=z,G_t=g]$ under the current policy, and
\cref{eq:upper-q} merely restricts the evaluation to boundary states.  Hence
\cref{eq:upper-v} collapses to
\begin{equation}
 V^{\mathrm{upper}}(z)
 =\E_{G\sim\mu_\eta(\cdot\mid z)}\bigl[V^{\mathrm{low}}(z,G)\bigr],
 \label{eq:upper-from-lower}
\end{equation}
so for a finite goal set one may simply define
\begin{equation}
 \widetilde V^{\mathrm{upper}}(z)
 :=\sum_{g\in\calG}\mu_\eta(g\mid z)\,V_\nu^{\mathrm{low}}(z,g)
 \label{eq:upper-critic-from-lower}
\end{equation}
use $\widetilde V^{\mathrm{upper}}$ in place of $V_\kappa^{\mathrm{upper}}$
in~\cref{eq:upper-critic}, and drop $\kappa$ altogether.  This removes a set of parameters and guarantees
that the two advantage estimators stay mutually consistent; it is the same
relation between the goal-conditioned and goal-marginal values that
option-critic methods exploit \citep{Bacon2017}.  The one caveat is a
training-distribution issue rather than a definitional one: boundary states are
not distributed like arbitrary states, so a single critic fit on all time steps
may be least accurate on precisely the sub-population where the upper-level
advantage estimate below evaluates it.

\paragraph{Advantage functions.}
The lower- and upper-level advantages, which we will use below
to estimate the actor's parameters, are given by
\begin{align}
 A^{\mathrm{low}}(z,g,a,d)
 &=Q^{\mathrm{low}}(z,g,a,d)-V^{\mathrm{low}}(z,g),
 \label{eq:lower-advantage}\\
 A^{\mathrm{upper}}(z,g)
 &=Q^{\mathrm{upper}}(z,g)-V^{\mathrm{upper}}(z).
 \label{eq:upper-advantage}
\end{align}
Each measures how much better a sampled lower-level decision or upper-level
goal is than the current policy's average decision at the same information
state.  For example, one-step actor--critic estimates are
\begin{align}
 \widehat A_t^{\mathrm{low}}
 &=R_{t+1}+\gamma V_\nu^{\mathrm{low}}(Z_{t+1},G_{t+1})
   -V_\nu^{\mathrm{low}}(Z_t,G_t),
 \label{eq:lower-advantage-estimate}\\
 \widehat A_k^{\mathrm{upper}}
 &=\mathcal R_k^{\mathrm{env}}+\gamma^{\Delta_k}
   V_\kappa^{\mathrm{upper}}(Z_{\tau_{k+1}})
   -V_\kappa^{\mathrm{upper}}(Z_{\tau_k}).
 \label{eq:upper-advantage-estimate}
\end{align}
These are exactly the regression residuals
of~\cref{eq:lower-critic,eq:upper-critic}, up to the target copies $\nu^-$ and
$\kappa^-$ used in the critic targets.  Actor and critic therefore share a
single temporal-difference computation at each level, and if one dispenses with
target networks they are literally the same quantity.

\paragraph{Fitting the actors.}
These estimators are used directly
to compute the policy gradient:
\begin{align}
 \nabla_\theta J
 &=\E\!\left[\sum_t
   \nabla_\theta\log\pi_\theta(A_t,D_t\mid Z_t,G_t)\,
   \widehat A_t^{\mathrm{low}}\right],
 \label{eq:lower-policy-gradient}\\
 \nabla_\eta J
 &=\E\!\left[\sum_k
   \nabla_\eta\log\mu_\eta(G_{\tau_k}\mid Z_{\tau_k})\,
   \widehat A_k^{\mathrm{upper}}\right].
 \label{eq:upper-policy-gradient}
\end{align}
Monte Carlo returns, multi-step returns, or generalized advantage estimation
can replace the one-step estimators.  If the actor objective uses the internal
reward, replace $R$ by $\bar R$ throughout---in the critic targets
of~\cref{eq:lower-critic,eq:upper-critic} as well as in the advantage
estimates, since an actor and a critic trained on different reward functions
make $\widehat A$ meaningless.  In that case $\mathcal R_k^{\mathrm{env}}$ is
replaced by the internal segment return $\mathcal R_k$ already defined
in~\cref{eq:smdp}.

\paragraph{Fitting the state update model.}
For a deterministic recurrent update, write $Z_t^\phi=U_\phi(h_t)$ for the
state obtained by unrolling the observation--action history $h_t$.  The return
gradient with respect to the state-update parameters is then estimated by
\begin{align}
 \nabla_\phi J
 &=\E\!\left[
   \sum_t \nabla_\phi
     \log\pi_\theta(A_t,D_t\mid Z_t^\phi,G_t)\,
     \widehat A_t^{\mathrm{low}}
   +\sum_k \nabla_\phi
     \log\mu_\eta(G_{\tau_k}\mid Z_{\tau_k}^\phi)\,
     \widehat A_k^{\mathrm{upper}}
 \right].
 \label{eq:state-policy-gradient}
\end{align}
The derivatives of the log policies include
$\nabla_\phi Z_t^\phi$ and are computed by backpropagation through time; the
environment itself need not be differentiated.  A stochastic $q_\phi$ instead
requires a reparameterization or score-function estimator.

\section{Conclusion}

In summary, the HHMM framing brings the following benefits:
\begin{enumerate}
  \item \textbf{Temporal commitment.}  The HHMM gate turns ``choose a goal''
  into a persistent intention rather than an independent per-step label.
  \item \textbf{Partial observability.}  Goal choice, action choice, and
  termination depend on the agent's filtered state $Z_t$, removing the need to
  identify observation with world state.
  \item \textbf{Call--return semantics.}  The done action becomes an explicit
  return from the within-goal controller to the goal selector.
  \item \textbf{Two valid objectives.}  Segment-level ACGVFs and continual
  hierarchical values differ exactly by the upper-level continuation term in
  \cref{eq:continual-bellman}.
  \item \textbf{Structured credit assignment.}  Primitive rewards train the
  lower controller, while boundary-to-boundary returns train the goal selector.
  \item \textbf{A route to deeper hierarchies.}  Additional states
  $Q_t^1,\ldots,Q_t^L$ and finish variables $F_t^1,\ldots,F_t^L$ represent
  nested intentions.  A level may change only when all descendants below it
  have returned, retaining the proper nesting constraint of an HHMM.
\end{enumerate}

In short, the ACGVF proposal supplies the missing agency---goals and completion
are chosen internally---while the recurrent-state agent supplies the missing
information-state machinery.  The HHMM supplies the missing temporal grammar:
select a goal, call a goal-conditioned controller, update internal state from
observations while the goal remains active, and return control upward when the
agent declares completion.

\bibliographystyle{plainnat}
\bibliography{bib}

\end{document}